\documentclass{article}

\usepackage{algorithm2e}

\usepackage{amsmath}
\usepackage{amssymb}

\usepackage{booktabs}
\usepackage{adjustbox}

\usepackage{hyperref}

\usepackage[accepted]{mlsys2022}

\usepackage{subfig}
\usepackage[]{graphicx}

\usepackage{todonotes}
\usepackage{multirow}
\usepackage{pgfplots}
\usepackage[section,frozencache,cachedir=.]{minted}

\newcommand{\STAB}[1]{\begin{tabular}{@{}c@{}}#1\end{tabular}} 

\pgfplotsset{small}

\usepackage[accepted]{mlsys2022}

\usepackage{algorithm2e}

\title{GPU Semiring Primitives for Sparse Neighborhood Methods}

\newcommand{\cuML}{RAFT}

\begin{document}


\twocolumn[
\mlsystitle{GPU Semiring Primitives for Sparse Neighborhood Methods}

\mlsyssetsymbol{equal}{*}

\begin{mlsysauthorlist}
\mlsysauthor{Corey J. Nolet}{nvidia,umbc}
\mlsysauthor{Divye Gala}{nvidia}
\mlsysauthor{Edward Raff}{umbc,booz}
\mlsysauthor{Joe Eaton}{nvidia}
\mlsysauthor{Brad Rees}{nvidia}
\mlsysauthor{John Zedlewski}{nvidia}
\mlsysauthor{Tim Oates}{umbc}
\end{mlsysauthorlist}

\mlsyscorrespondingauthor{Corey J. Nolet}{cjnolet@gmail.com}

\mlsysaffiliation{nvidia}{NVIDIA}
\mlsysaffiliation{booz}{Booz Allen Hamilton}
\mlsysaffiliation{umbc}{University of Maryland, Baltimore County}

\mlsyskeywords{Sparse Distances, GPUs}

\vskip 0.3in

\begin{abstract}

High-performance primitives for mathematical operations on sparse vectors must deal with the challenges of skewed degree distributions and limits on memory consumption that are typically not issues in dense operations. We demonstrate that a sparse semiring primitive can be flexible enough to support a wide range of critical distance measures while maintaining performance and memory efficiency on the GPU. We further show that this primitive is a foundational component for enabling many neighborhood-based information retrieval and machine learning algorithms to accept sparse input. To our knowledge, this is the first work aiming to unify the computation of several critical distance measures on the GPU under a single flexible design paradigm and we hope that it provides a good baseline for future research in this area. Our implementation is fully open source and publicly available as part of the RAFT library of GPU-accelerated machine learning primitives (https://github.com/rapidsai/raft).
\end{abstract}
]

\printAffiliationsAndNotice{}  

\section{Introduction}

Many machine learning and information retrieval tasks operate on sparse, high-dimensional vectors. Nearest-neighbor based queries and algorithms in particular are instrumental to many common classification, retrieval, and visualization applications\cite{scholkopf2001generalized,alpay2012reproducing,berlinet2011reproducing,smola2007hilbert,scholkopf2018learning}. As General-purpose GPU computing (GPGPU) has become more popular, the tools for IR and distance computations on GPUs has not kept pace with other tooling on dense representations like image and signal processing that have contiguous access patterns that are easier to code for \cite{guo2020accelerating}. Sparse methods of linear algebra on GPUs have long existed, though they are often specialized and difficult to adapt to new distance measures. This stems from having to account for various hardware and application-specific constraints \cite{jeon2020biqgemm,guo2020accelerating,gale2020sparse,gray2017gpu, bell2008efficient}, and assumptions on the distribution of non-zeros in the input and output data \cite{sedaghati2015characterizing,mattson2013standards}. 
This complexity and specialization has slowed the adoption for sparse data and operations in general purpose tools like PyTorch and Tensorflow.

To develop a more general code base that supports good performance and flexibility for new distance measures on sparse data, we develop an approach leveraging Semirings. 
Semirings provide a useful paradigm for defining and computing inner product spaces in linear algebra using two operations, as in the MapReduce \cite{mattson2013standards,emoto2012filter} paradigm, where a \textit{product()} function is used to define a mapping between point-wise corresponding elements of vectors and a \textit{sum()} function is used to reduce the products into a scalar. Using semirings to implement algorithms with sparse linear algebra on GPUs is an active area of research \cite{fender2017parallel, 9150461, lettich2021galatic} and has been widely studied for helping to consolidate both the representation and execution of operations on graphs and probabilistic graphical models. In this paper, we show that semirings can be used for sparse neighborhood methods in machine learning, extending the benefits to all algorithms capable of using them.
We define semirings more formally in \autoref{subsec:semirings} but use the more general description above to navigate the benefits and related work in the following section.


A common issue for large-scale sparse problems in high-performance single-instruction multiple-data (SIMD) environments, like the GPU, is load balancing in order to keep the processing units constantly moving forward. As we will show in Section \ref{sec:gpu_acceleration}, the imbalanced load and resource requirements for a simple and straightforward naive semiring implementation, capable of computing distances like Manhattan, suffers from large thread divergences within warps, highly uncoalesced global memory accesses, and resource requirements which are unrealistic in many real-world datasets.

In order to integrate into an end-to-end data science or scientific computing workflow, such as in the PyData or RAPIDS~\cite{raschka2020machine} ecosystems, an efficient implementation of a primitive for computing pairwise distances on sparse datasets should ideally preserve as many of the following characteristics as possible. In this paper, we show that our implementation preserves more of the below characteristics than any other known implementation.

\begin{enumerate}
    \item Maintain uniformity of intra-warp instruction processing.
    \item Coalesce both reads from and writes to global memory.
    \item Process data inputs without transposition or copying.
    \item Use as little memory as necessary.
    \item Enable semirings in addition to the simple dot product.
\end{enumerate}

\begin{table*}[ht]
\caption{\label{tab:semiring_distances} Common distances and their semirings. While all distances can be computed with the NAMM (where $id_\otimes = 0$), the distances in this table which require it have their $\otimes$ listed. The expansion function and any potential norms are provided for distances that can be computed in the more efficient expanded form.}
\centering
\bgroup
\def\arraystretch{1.8}
\begin{tabular}{|| lcccc ||}
 \hline
 Distance & Formula & NAMM & Norm & Expansion\\ [0.5ex] 
 \hline\hline 
 Correlation & $1 - \frac{\sum_{i=0}^k{(x_i - \bar{x})(y_i - \bar{y})}}{\sqrt{\sum_{i=0}^k{x_i - \bar{x}^2}}^2 \sqrt{\sum_{i=0}^2{y_i - \bar{y}^2}}^2}$ & & $L_1$,$L_2$ & $1 - \frac{k \langle x \cdot y\rangle  - \|x\|\|y\|}{\sqrt{(k \|x\|_2 - \|x\|^2) (k\|y\|_2 - \|y\|^2)}} $ \\

\hline 
 Cosine & $\frac{\sum_{i=0}^k{x_iy_i}}{\sqrt{\sum_{i=0}^k{x_i^2}}\sqrt{\sum_{i=0}^k{y_i^2}}}$ & & $L_2$ & $1 - \frac{\langle x \cdot y\rangle }{\|x\|_2^2\|y\|_2^2}$ \\
 \hline
Dice-Sorensen & $\frac{2|\sum_{i=0}^k{x_iy_i}|}{(\sum_{i=0}^k{x})^2 + (\sum_{i=0}^k{y})^2}$ & & $L_0$ & $\frac{2\langle x \cdot y\rangle }{|x|^2 + |y|^2 }$ \\
\hline
 Dot Product & $\sum_{i=0}^k{x_i y_i}$ & & &$\langle x \cdot y\rangle $ \\ 
 \hline
 Euclidean & $\sqrt{\sum_{i=0}^k{|x_i-y_i|^2}}$ & & $L_2$ & $\|x\|_2^2 - 2\langle x \cdot y\rangle  + \|y\|_2^2$ \\
  \hline
Canberra & $\sum_{i=0}^k{\frac{|x_i-y_i|}{|x_i| + |y_i|}}$  & $\{\frac{|x-y|}{|x| + |y|}, 0\}$ & & \\
  \hline
Chebyshev & $\sum_{i=0}^k{\max(x_i-y_i)}$ & $\{\max(x-y), 0\}$ & & \\
 \hline
 
Hamming & $\frac{\sum_{i=0}^k{x_i \neq y_i}}{k}$ & $\{x\neq y, 0\}$ & &\\
\hline
Hellinger & $\frac{1}{\sqrt{2}}\sqrt{\sum_{i=0}^k{(\sqrt{x_i} - \sqrt{y_i})^2}}$ & & & $1 - \sqrt{\langle \sqrt{x} \cdot \sqrt{y}\rangle }$ \\
\hline

Jaccard & $\frac{\sum_{i=0}^k{x_iy_i}}{(\sum_{i=0}^k{x_i^2}+\sum_{i=0}^k{y_i^2} - \sum_{i=0}^k{x_iy_i}}$ & & $L_0$ & $1 - \frac{\langle x \cdot y\rangle }{(\|x\| + \|y\| - \langle x \cdot y\rangle )}$ \\
\hline
Jensen-Shannon & $\sqrt{\frac{\sum_{i=0}^k{x_i\log{\frac{x_i}{\mu_i} + y_i\log{\frac{y_i}{\mu_i}}}}}{2}}$ & $\{x\log{\frac{x}{\mu}} + y\log{\frac{y}{\mu}}, 0\}$ & & \\
\hline
KL-Divergence & $\sum_{i=0}^k{x_i \log(\frac{x_i}{y_i})}$ &  & & $\langle x \cdot \log{\frac{x}{y}}\rangle$  \\
\hline
 Manhattan & $\sum_{i=0}^k{|x_i-y_i|}$ & $\{|x-y|, 0\}$ & &  \\
 \hline
 Minkowski & $(\sum_{i=0}^k{|x_i-y_i|^p})^{1/p}$ & $\{|x-y|^p, 0\}$ & & \\ 
\hline
Russel-Rao & $\frac{k - \sum_{i=0}^2{x_i y_i}}{k}$ & &  & $\frac{k - \langle x \cdot y\rangle }{k}$  \\

\hline

\end{tabular}
\egroup
\end{table*}

\section{Semirings and Pairwise Distances} 
\label{sec:semirings_pairwise_dists}


We formalize the concepts of semirings and distance measures in this section and describe building blocks required to implement several popular distance measures, often encountered in machine learning applications, into the semiring framework. 

In machine learning applications, a distance measure is often performed on two row matrices containing data samples with columns that represent some number of observations, or features. In this paper, we will refer to these two matrices as $A$ and $B$ in upper-case where $A\in\mathbb{R}^{m\times k}$, and $B\in\mathbb{R}^{n\times k}$ and a single vector as $a$ and $b$ in lowercase where $a\in\mathbb{R}^k$ or $b\in\mathbb{R}^k$. As we show in this section, the core of computing pairwise distances between $A$ and $B$ is a matrix multiplication $A B^\top$ in a topological space equipped with an inner product semiring that defines distances between vectors. When this inner product is defined to be the dot product semiring, the topological space defines the standard matrix multiply but we can capture many other core distances in machine learning applications by simply redefining the inner product semiring.

While some distance measures can make use of the simple dot product semiring from matrix-matrix multiplication routines, we show that a more comprehensive package for computing pairwise distances requires more flexibility in terms of the arithmetic operations supported. Further, the explicit transposition of $B$ which is required in routines such as the cuSPARSE \textit{csrgemm()} requires a full copy of $B$, since no elements can be shared between the original and transposed versions in the CSR data format. This has a negative impact on scalability in memory-constrained environments such as GPUs.

\subsection{Distances}
\label{sec:distances}

Sparse matrix-matrix multiplication with a standard dot product semiring is most performant in cases where only the intersection is needed between pairs of corresponding nonzero columns in each vector. Because a standard multiplication between two terms has an identity of 1 and multiplicative annihilation (e.g. $a_i * 0 = 0$), the dot product semiring between two vectors can be computed efficiently by iterating over the nonzero columns of one vector and only computing the product of the corresponding nonzero columns of the other vector. Many distances can make use of this property, 
in table \ref{tab:semiring_distances} we dervice the semi-ring annihilators and expansions (as needed) for 15 distances. 





    

For a distance to define a metric space, it must follow four properties- implication ($d(a,b)=0 \implies a=b$), positivity ($d(a,b)>=0)$, symmetry ($d(a,b)=d(b,a)$), and the triangle inequality ($ d(a,c) \le d(a,b) + d(b, c)$). Several metrics, including Chebyshev, Manhattan, and Euclidean, are derived from the generalized Minkowski formula $\left(\sum_i^k{|a_i-b_i|^{p}}\right)^{1/p}$ where $p$ defines a degree.  The absolute value in this equation defines a commutative semiring which requires commutativity in the difference of each vector dimension. Euclidean distance is equivalent to Minkowski with a degree of 2 ($(\sum_i^k{|a_i-b_i|^2})^{1/2}$). Because the square of a number is always positive, this equation can be expanded to $(a - b)^p$ for all even degrees and still preserve the absolute value, such as $(a-b)^2 = a^2 - 2\langle a, b\rangle + b^2$ in the case of Euclidean distance. While numerical instabilities can arise from cancellations in these expanded equations, we will show in section \ref{subsec:semirings} that the expanded form is often preferred in sparse algebras, when distances can make use of it, because it requires less computations than the exhaustive evaluation over the nonzeros of $k$. By example, the distances which don't have an expanded form, such as Manhattan (Minkowski with degree 1) and Chebyshev (Minkowski with degree $max$) distance, are often non-annihilating (e.g. $x * 0 = x$) and require computation over the full union of nonzero columns from both vectors in order to preserve commutativity.

\subsection{Semirings}
\label{subsec:semirings}
A \textit{monoid} is a semigroup containing an associative binary relation, such as addition ($\oplus$), and an identity element ($id_\oplus$). A \textit{semiring} \cite{ratti1971graphs}, denoted $(S, \mathbb{R}, \{\oplus, id_\oplus\}, \{\otimes, id_\otimes\})$, is a tuple endowed with a domain along with additive ($\oplus$) and multiplicative ($\otimes$) monoids where 

\begin{enumerate}
    \item $\oplus$ is commutative, distributive, and has an identity element 0
    \item $\otimes$ distributes over $\oplus$
\end{enumerate}
 
Some formal definitions of semirings require that $id_\otimes=1$. Given two sparse vectors $a, b \in \mathbb{R}^k$, a semiring with $(S, \mathbb{R}, \{\oplus, 0\}, \{\otimes, 1\})$ and  $annihilator_\otimes=0$ has the effect of only requiring $\otimes$ be computed on columns that are both nonzero (e.g. $nonzeros(a) \cap nonzeros(b))$. These rules are often relaxed in practice, for example in tropical semirings in \autoref{eq:tropical_semiring}, which can solve dynamic programming problems such as the Viterbi algorithm. An \textit{annihilator} is an input that will always cause a monoid to evaluate to 0 and the multiplicative annihilator ($annihilator_\otimes$)  is often assumed to be $id_\oplus$. A monoid is non-annihilating when it does not have a defined annihilator. When an expanded form is not possible or efficient, $\otimes$ also must be commutative in metric spaces, and thus must be non-annihilating and $id_\otimes = 0$. We refer to this monoid as a \textit{non-annihilating multiplicative monoid} (NAMM).

\begin{equation}
\label{eq:tropical_semiring}
    (S, \mathbb{R} \cup \{+ \infty\}, \{min, +\infty\}, \{+, 0\})
\end{equation}
 



Table \ref{tab:semiring_distances} uses semirings to construct several commonly used distances common in machine learning and data science applications. When an expanded form is possible, an expansion function can be performed as an element-wise operation on a simple pairwise dot product semiring with arrays of row-vector norms. While most of the expanded form distances can directly use the dot product semiring, KL-divergence directly replaces the $\otimes$ with $a_i \log(a_i/b_i)$ and makes no further assumption of symmetry. A NAMM is required for all unexpanded distance measures where $id_\otimes = 0$ and special care must be taken to ensure it is applied to the full union of the non-zero columns of corresponding elements from each pair of vectors. 

As mentioned in the previous section, the Euclidean distance can be expanded to $\|A\| - 2\langle A B^\top\rangle + \|B\|$. This equation can be decomposed into the sum of individual L2 norms, a matrix product, and an element-wise expansion function executed in parallel over the individual dot products from the matrix product to combine the parts into a single scalar distance. Given vectors $A_i, B_j$, the expansion function for Euclidean distance can be derived by distributing their squared difference over the exponent to produce $(A_i-B_i)\times(A_i-B_i)$ and further expanding it to $\|A\|_i + 2 \langle A_i, B_j\rangle - \|B\|_j$.

The $annihilator_\otimes$ and $id_\otimes$ determine the number of times the $\otimes$ monoid must be applied during the computation of pairwise distances. When $annihilator_\otimes = id_\oplus$, then $\otimes(a_i, 0) = 0$ and $\otimes(0, b_i) = 0$ so $\otimes$ can be applied only to the intersection of columns. When $annihilator_\otimes$ is undefined and $id_\otimes = 0$, then $\otimes$ must be applied exhaustively over the union of columns because $\otimes(a_i, 0) = a_i$ and $\otimes(0, b_i) = b_i$.

A union between two sets can be decomposed into an intersection between the two sets, along with the union of the symmetric differences between them. These are shown in \autoref{eq:full_union}, where a complement is denoted with a $\overline{bar}$. The nonzero columns of two sparse vectors can be used as sets $a$ and $b$ in this equation and the sparse matrix multiply with an ordinary dot product only requires the application of \textit{product()} across $a \cap b$. The NAMM, however, requires the application of the \textit{product()} across the full union of nonzero columns $a \cup b$.


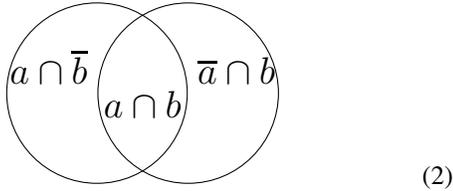
  \def\secondcircle{(210:0.7) circle (1.2)}
  \def\thirdcircle{(330:0.7) circle (1.2)}
  \begin{align}
    \begin{tikzpicture}
      \begin{scope}
    \clip \secondcircle;
      \end{scope}
      \begin{scope}
      \end{scope}
        \draw \secondcircle node [text=black, above left] {\Large $a \cap \overline{b}$};
      \draw \thirdcircle node [text=black,above right] {\Large $\overline{a} \cap b$};
    \node at (0, -0.5) [text=black] {\Large $a \cap b$};
    \end{tikzpicture}
    \end{align}

\begin{align}
\label{eq:full_union}
a \cup b =  \{a \cap b \} \cup \{ \overline{a} \cap b \} \cup \{ a \cap \overline{b}\}
\end{align}

A common approach to implementing sparse matrix multiply is to iterate over the nonzeros from $b$ in order to lookup and compute the intersection with the nonzeros from $a$. This design will also implicitly compute the symmetric difference between either of the two sets of nonzeros, $a\cap \overline{b}$ or $\overline{a}\cap b$, depending on which vector is chosen in the iteration over nonzeros. To compute a full union, the remaining set difference can be computed in a second pass of the matrix multiply by looping over the nonzeros from the vector that remains. We will show in \autoref{sec:gpu_acceleration} that we accomplish this efficiently in our implementation in two passes- one pass to compute the first two terms and another pass to compute the third term. Distances which can be computed with an expansion function only need the first pass while distances which require the NAMM need both. Please refer to \autoref{deriving_distances_with_semirings} for an example of using semirings to compute the Manhattan distance using the NAMM.

Existing semiring implementations currently require that the $id_\oplus$ be used as $annihilator_\otimes$. For example, the GraphBLAS specification enables the re-interpretation of the zeroth element, but this is necessary to define the identity of the $\oplus$ monoid.

\section{GPU-Accelerated Semirings}

In this section, we briefly introduce GPU architecture before discussing some naive designs and the inefficiencies that led to the construction of our final design. Our goal was to preserve as many of the ideal design characteristics from \autoref{sec:related_work} as possible but we found a need to accept trade offs during implementation. 


\subsection{GPU Architecture}
\label{sec:gpu_acceleration}

The largest GPUs today contain hundreds of hardware processing cores called streaming multiprocessors (SM) which execute groups of threads called warps. Each warp can process a single instruction at a time in parallel using a paradigm called single-instruction multiple data (SIMD). It's important that threads within a warp minimize conditional branching that will cause the threads to wait for each branch to complete before proceeding. This is called thread divergence, and can severely limit effective parallel execution. On the Volta and Ampere architectures, each SM can track the progress of up to 64 warps concurrently \cite{tesla2018v100}, and rapidly switch between them to fully utilize the SM. Each SM has a set of registers available which allows warps to perform collective operations, such as reductions. Warps can be grouped into blocks and a small amount of memory can be shared across the threads and warps. 

Global, or device, memory can be accessed by all of the SMs in the GPU. Accesses to contiguous device memory locations within a warp can be coalesced into a single blocked transaction so long as the accesses are performed in the same operation. In SIMD architectures, uniform patterns can be critical to performance unless latencies from non-uniform processing, such as uncoalesced memory accesses, can be hidden with increased parallelism.

Registers provide the fastest storage, and it's generally preferable to perform reductions and arithmetic as intra-warp collective operations where possible. Intra-block shared memory is also generally preferred over global memory when a problem can be made small enough to benefit. However, contiguous locations of shared memory are partitioned across contiguous banks and any accesses to different addresses in the same bank by the same warp will create a bank conflict and be serialized within the warp, causing the threads to diverge.

\subsection{Naive Semi-Ring Full-Union CSR Designs}
\label{sec:naive_semiring}

\subsubsection{Expand-Sort-Contract}
Initial implementations tried to minimize the memory footprint as much as possible by directly computing the output distances from the input CSR format. The CSR format requires columns to be sorted with respect to row and we initially attempted to use a modified variant of the \textit{expand-sort-contract}~\cite{dalton2015optimizing} pattern on the nonzero columns from each pair of row vectors, $a, b \in \mathbb{R}^k$, concatenating the vectors together, sorting them, and applying the $\otimes$ monoid on pairs of duplicate columns to \textit{contract} the sorted array and invoking $\otimes$ with the identity for all other columns. At the row-level of the output matrix, no computations would be able to be reused by subsequent pairs of vectors so we implemented this pattern on the GPU and mapped the nonzero columns and values for each row-vector pair to individual thread-blocks, \textit{expanding} both vectors by concatenating them in shared memory, performing a sort-by-key, and compressing them in parallel. We attempted several efficient sorting algorithms on the GPU including the popular radix sort and bitonic sorting networks and, while the use of shared memory in the sort step enabled coalesced reads from global memory for the nonzero columns and values, the sorting step dominated the performance of the algorithm. Another downside with this particular design is that both vectors need to fit in shared memory, requiring space for $2 * (nonzeros(a) + nonzeros(b))$ elements in order to fit both the columns and corresponding values at the same time. In addition to the need for $n*m$ blocks to be scheduled, the shared memory requirement became a severe limit to scale, which was further compounded by the shared memory size limiting the number of blocks that could be scheduled concurrently on each SM.

\begin{algorithm}
\SetAlgoLined
\KwIn{$A_i, B_j, product\_op, reduce\_op$}
\KwResult{$C_{ij} = d(A_i, B_j)$}
 $smem[0..nnz_{a_{i-1}}] = A_i$\;
 
 $smem[nnz_{a_i}..nnz_{b_{j-1}}] = B_j$\;
 
 $sort(smem)$\;
 
 $C_{ij}$ = reduce(smem, $product\_op$, $reduce\_op$)\;
 \caption{Semiring on CSR inputs using expand-sort-contract pattern, parallelized across threads in each block.} \label{alg:naive_semiring_sort}
\end{algorithm}

\subsubsection{Iterating Sorted Nonzeros}
\label{sec:iterating_sorted_nonzeros}
Since columns will often be sorted within their respective rows in the CSR format, we removed the sort step from \autoref{alg:naive_semiring_sort} by exhaustively iterating over the non-zeros of each $O(m*n)$ pair of vectors in parallel, one pair per thread, as shown in \autoref{alg:naive_semiring_nosort}.  We found that even when the neighboring threads processed rows of similar degree, the differing distributions of nonzeros within each row decreased the potential for coalesced global memory accesses and created large thread divergences. Further, the exhaustive nature of this design, while it will guarantee the $\otimes$ monoid is computed on the full union of nonzero columns, will end up performing many unnecessary computations when distances can be computed with the rules of a simple dot product semiring.

\begin{algorithm}
\SetAlgoLined
\KwIn{$A_i, B_j, product\_op, reduce\_op$}
\KwResult{$C_{ij} = d(A_i, B_j)$}
 startA = $indptrA_i$, endA = $indptrA_{i+1}$\;
 
 startB = $indptrB_j$, endB = $indptrB_{j+1}$\;
 
 $i_{colA}$ = startA, $i_{colB}$ = startB\;
 
 \While{$i_{colA}$ $<$ endA || $i_{colB}$ $<$ endB}{
 
  colA = $i_{colA}$ $<$ endA ? $indices_{i_{colA}}$ : MAX\_INT\;
  
  colB = i\_colB $<$ endB ? $indices_{i_{colB}}$ : MAX\_INT\;
  
  valueA = 0, valueB = 0\;
  \If{colA $\leq$ colB}{
  
    valueA = $valuesA_{i_{colA}++}$\;
   }
  \If{colB $\leq$ colA}{
  
    valueB = $valuesB_{i_{colB}++}$\;
   }
   
  $v = product\_op(valueA, valueB)$\;
  
  $C_{ij} = reduce\_op(C_{ij}, v)$\;
 }
 \caption{Semring on CSR inputs. Each thread computes a single dot product.} \label{alg:naive_semiring_nosort}

\end{algorithm}

We found marginal gains in performance by coalescing the reads of the vectors from $A$ into shared memory and sharing it across all threads of each thread-block. We attempted to load balance this algorithm by maintaining arrays to look up row information for each column but this increased warp divergence from the overly complicated conditionals required to maintain state across threads and warp boundaries.

\subsection{Load Balanced Hybrid CSR+COO}
\label{sec:load_balanced_semiring}

While the CSR format enables algorithms to be parallelized over threads for individual rows, we found that using a row index array in coordinate format (COO) for $B$ enabled load balancing, coalescing the loads from each vector from A into shared memory, once per block, and threads of each block parallelizing the application of the semiring over nonzero elements of B. Since the columns in B are assumed to be sorted by their respective row, we use a segmented reduction by key within each warp, bounding the number of potential writes to global memory by the number of active warps over each row of $B$. Our design extends the work of the COO sparse-matrix dense-vector multiplication described in \cite{Anzt2020} by storing the vectors from $A$ in dense form in shared memory only when the number of columns are small enough. Our extension enables sparse-matrix sparse-vector multiplication by storing the vectors in sparse form when their degrees are small enough. We achieve full occupancy on the Volta architecture by trading off the size of the L1 cache to double the amount of shared memory per GPU, allowing each SM to use 96KiB. Since our design uses less than 32 registers, a block size of 32 warps allows two blocks, the full 64 warps, to be scheduled concurrently on each SM.

\begin{algorithm}
\SetAlgoLined
\SetKw{KwBy}{by}
\KwIn{$A_i, B, product\_op, reduce\_op$}
\KwResult{$C_{ij} = d(A_i, B_j)$}
read $A_i$ into shared memory\;

cur\_row=rowidx[ind]\; 

ind = idx of first elem to be processed by this thread\;

c = product\_op(A[ind], x[colidx[ind]])\;

\For{$i\gets1$ \KwTo $\mathit{nz\_per\_chunk}$; \KwBy $\mathit{warp\_size}$}
{
    next\_row = cur\_row + $\mathit{warp\_size}$\;
    
    \If{next\_row != cur\_row || $\mathit{is\_final\_iter?}$}
    {
        v = segmented\_scan(cur\_row, c, product\_op)\;
        
        \If{$\mathit{is\_segment\_leader?}$}
        {
            atomic\_reduce(v, reduce\_op)\;
        }
        c = 0\;
    }
    
    cur\_row = next\_row\;
    
    ind += $warp\_size$\;
    
    c = product\_op(A[ind], x[colidx[ind]])\;
}
 \caption{Load-balanced Hybrid CSR+COO SPMV.} \label{alg:naive_semiring_coo_spmv}

\end{algorithm}

\subsubsection{Two-pass execution}

As described in \autoref{subsec:semirings}, a single execution of this strategy will compute the intersection and symmetric difference $\overline{a}\cap b$ between nonzero columns from each vector $a$, and $b$ so long as $\otimes$ is applied to all nonzero columns of $b$. While only a single pass covers distance measures which require only a column intersection (e.g. dot product semiring $(S, \mathbb{R}, \{+, 0\}, \{*, 1\})$), a second pass can compute the remaining symmetric difference required for the full union between non-zero columns by commuting $A$ and $B$ and skipping the application of of $id_\otimes$ in $B$ for the second pass.

\subsubsection{Sparsifying the Vector in Shared Memory}

While we found storing the vectors from $A$ in dense form in shared memory to have the highest throughput rate and least amount of thread divergence within each warp, sparse datasets are generally assumed to have high dimensionality and the limited amount of shared memory that can be allocated per SM bounds the size of the vectors that can be stored in it. For example, The $96KiB$ limit per block on Volta allows a max dimensionality of $~23K$ with single-precision and the $163KiB$ limit per SM on Ampere allows a max dimensionality of $~40K$ with single-precision. Coupling the amount of shared memory to the dimensionality creates a problem for occupancy as it approaches capacity. Both of these architectures limit the maximum block sizes to 1024 threads and max concurrent warps per SM to 64 so anything over $48KB$ of shared memory per block is going to decrease occupancy. For this reason, the maximum dimensionality of dense vectors that can be processed with full occupancy is actually $~12K$ and $~20K$, respectively. 

This boundary becomes too small for many sparse datasets which would instead benefit from coupling the shared memory size to individual row degrees. Inspired by other sparse matrix multiplication implementations on the GPU \cite{anh2016balanced, kunchum2017improving, liu2014efficient,nagasaka2017high}, we enhanced the vector insertion and lookup patterns of the COO SPMV design outlined in \cite{Anzt2020} by building a hash table to store these columns in shared memory. Unlike many other hash table implementations on the GPU \cite{alcantara2009real, ashkiani2018dynamic, alcantara2012building, pan2011fast,cassee2017analysing}, our implementation builds an independent hash table per thread-block and so many other designs and concurrency patterns that optimize the key distribution and collision-resolution strategies for the GPU are not efficient or cannot be easily ported for our use-case. For this reason, we used a simple hash table with a \textit{Murmur} hash function and linear probing and leave the investigation of a better and more optimized design to future work.

Hash tables have the best performance when the number of entries is less than 50\% of the capacity. As the hash table size grows beyond 50\% capacity, the collision resolution cycles of linear probing, which are non-uniform, increase the serialization of instructions from warp divergences and also increase the number of transactions from global memory reads of $B$ since they can no longer be coalesced. The hash table strategy decreases the amount of shared memory available, often by a factor of 2, because the nonzeros need to be stored together as key/value pairs to avoid an additional costly lookup to global memory, a side-effect which would only further increase serialized execution from diverging threads. Our hash table strategy allows for a max degree of $~3K$ on Volta architectures and $~5K$ on Ampere.

Another unfortunate side-effect from the linear-probing collision strategy of our hash table is the increase in lookup times for columns even for elements that aren't in the table. For example, as the hash table approaches capacity, the increase in collisions can cause a lookup to probe through multiple candidates, sometimes hundreds, before finding an element doesn't exist. Bloom filters have been used to implement fast list intersection problems for sparse matrix multiplication problems on the GPU \cite{zhang2020sparch, zhang2011fast}. As an alternative to the hash table approach, we tried building a bloom filter in shared memory and used a binary search to perform lookups of nonzeros in global memory for positive hits. While we found this technique to yield marginally better performance on the Jensen-Shannon distance in one of our benchmarks, likely because it helped hide some of the compute-bound latencies from the additional arithmetic, we were not able to extract a simple rule from the data shapes or sparsity patterns that would allow us to know, before starting the computation, when it should be used.


\subsubsection{Handling High Degree Columns}

Our hash table implementation shows reasonable performance up to 50\% capacity. Rows with degree greater than 50\% hash table capacity are partitioned uniformly by their degrees into multiple blocks with subsets of the degrees that can fit into 50\% hash table capacity. Using a similar logic to that of blocked sparse techniques, our partitioning strategy does extra work in exchange for scale. Further, this technique requires each thread perform a branching conditional so it can test whether each nonzero column of $B$ is part of the current partition. As we show in \autoref{sec:experiments}, we do find that this strategy can perform well on some datasets when most of the degrees are small enough to fit in the hash table. For example, we found this strategy spent a miniscule amount of time in this step on the Movielens dataset.



\subsection{Norms and Expansion Functions}
\label{sec:additional_building_blocks}

Distances which can be computed in their expanded forms can use the dot product semiring directly and only require a single pass through our SPSV. Computing distances in their expanded form often requires one or more vectors of row norms as well as an \textit{expansion function}, which uses some arithmetic to combine the norm vectors with the individual dot products (refer to \autoref{tab:semiring_distances} for examples). Row norms can be computed over CSR matrices using a row-wise reduction on the GPU as each row can be mapped to a single block or warp and the norm computed by a warp-level collective reduction. The reduction primitive necessary for computing these row norms is already part of the GraphBLAS specification.

The actual arithmetic in each expansion function is dependent upon the distance measure, however the kernel to apply the expansion function can be executed embarrassingly parallel using an element-wise primitive, also part of the GraphBLAS specification, to map each entry in the dot product matrix to an individual GPU thread to coalesce the reads and writes.

\section{Experiments}
\label{sec:experiments}
We evaluated the runtime performance characteristics and generalization of our approach by benchmarking our semiring strategies against several real-world sparse datasets with different shapes and degree distributions. We also analyze the GPU memory footprint of the cuSPARSE \textit{csrgemm()} and our load-balanced COO SPMV.

\subsection{Datasets}

The datasets which we found are often used to benchmark sparse matrix-matrix and matrix-vector implementations on the GPU demonstrate the subtle differences in the objectives between using semirings for sparse neighborhood methods and using sparse linear algebra more generally for things like graph algorithms and eigendecompositions. As an example, one such set of datasets which we found commonly used in papers to benchmark sparse linear algebra implementations \cite{williams2007optimization,bell2008efficient} is composed almost entirely of square connectivities graphs, and these would not provide a useful performance indicator for the objective of creating connectivites graphs from bipartite graphs. For this reason, and the lack of prior research in our objective, we establish a new baseline using datasets that our algorithm would be expected to encounter in practice. Our baseline uses cuSPARSE for all the expanded distance measures, along with the naive CSR full-union semiring implementation as described in section \ref{sec:iterating_sorted_nonzeros} for the distances which cuSPARSE does not support.

The \textit{MovieLens}~\cite{harper2015movielens} Large dataset contains ratings given by 283k users for 194k movies. We used a dataset of 70k cells and gene expressions for 26k genes from the human cell atlas \cite{travaglini2020molecular} as an example of a single-cell RNA workflow. For natural language processing examples, we benchmarked two different datasets containing TF-IDF vectors for two different use-cases. We used the NY Times Bag of Words dataset\cite{newmann2008} for an example of document similarity and n-grams generated from a list of company names from the SEC EDGAR company names database for an example of string matching.

\begin{table}[!h] 
\caption{Datasets used in experiments} \label{tab:datasets}
\adjustbox{max width=\columnwidth}{%
\begin{tabular}{lrrrr}
\toprule
Dataset & Size & Density & Min Deg & Max Deg\\ \midrule
Movielens Large                 & (283K, 194K)  & 0.05\% & 0 & 24K \\ 
SEC Edgar       & (663K, 858K) & 0.0007\% & 0 & 51  \\
scRNA              & (66K, 26K) & 7\% & 501 & 9.6K  \\
NY Times BoW & (300K, 102K) & 0.2\% & 0 & 2K \\
\bottomrule
\end{tabular}
}
\end{table}
\begin{figure}
\centering
\begin{tikzpicture}
\begin{axis}[
    legend style={at={(0.5,-0.2)},anchor=north},
    legend columns=4, 
    height=0.6\columnwidth,
    width=\columnwidth,
    xmode=log,
    xmin=1
]
\addplot+ table [x=x, y=y, col sep=comma,mark=none] {degree_cdf/nytimes_degree_cdf.csv};
\addlegendentry{ny times}
\addplot+ table [x=x, y=y, col sep=comma,mark=none] {degree_cdf/movielens_degree_cdf.csv};
\addlegendentry{movielens}
\addplot+ table [x=x, y=y, col sep=comma,mark=none] {degree_cdf/scrna_degree_cdf.csv};
\addlegendentry{scrna}
\addplot+ table [x=x, y=y, col sep=comma,mark=none] {degree_cdf/string_matching_degree_cdf.csv};
\addlegendentry{sec edgar}
\end{axis}
\end{tikzpicture}



\caption{CDFs of Degree Distributions for the datasets used in our benchmark on the interval 0-99\%. We can see that 99\% of the degrees in the SEC Edgar datasets are <10 while 88\% of the degrees for Movielens are <200. On average scRNA has the largest degrees with 98\% of the rows having degree 5k or less. The NY Times dataset has the highest variance, with 99\% of the rows having degree less than 1k.}
\end{figure}
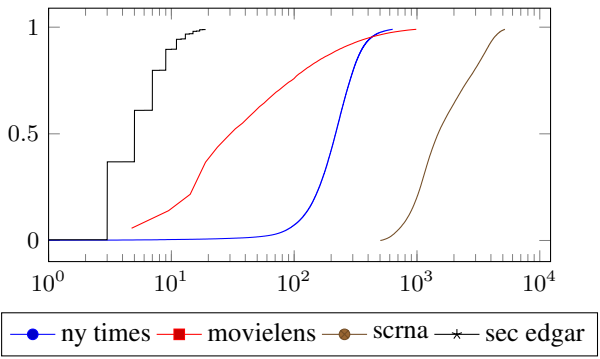

\begin{table*}
\caption{Benchmark Results for all datasets under consideration. All times are in seconds, best result in \textbf{bold}. The first italicized set of distances can all be computed as dot products, which are already highly optimized for sparse comparisons today. This easier case we are still competitive, and sometimes faster, than the dot-product based metrics. The Non-trivial set of distances that are not well supported by existing software are below, and our approach dominates  amongst all these metrics.} \label{tbl:gpu_results_movielens} \label{tbl:gpu_results_nytimes} \label{tbl:gpu_results_scrna} \label{tbl:gpu_results_secedgar}
\begin{tabular}{@{}llrrrrrrrr@{}}
\toprule
                                                 & \multicolumn{1}{c}{} & \multicolumn{2}{c}{MovieLens}                            & \multicolumn{2}{c}{scRNA}                                & \multicolumn{2}{c}{NY Times Bag of Words}                & \multicolumn{2}{c}{SEC Edgar}                            \\ \cmidrule(l){3-4}  \cmidrule(l){5-6} \cmidrule(l){7-8} \cmidrule(l){9-10}
\multicolumn{2}{l}{Distance}                                            & \multicolumn{1}{c}{Baseline} & \multicolumn{1}{c}{\cuML} & \multicolumn{1}{c}{Baseline} & \multicolumn{1}{c}{\cuML} & \multicolumn{1}{c}{Baseline} & \multicolumn{1}{c}{\cuML} & \multicolumn{1}{c}{Baseline} & \multicolumn{1}{c}{\cuML} \\ \midrule
\multirow{7}{*}{\STAB{\rotatebox[origin=c]{90}{Dot Product Based}}}   & \textit{Correlation} & 130.57                       & \textbf{111.20}           & \textbf{207.00}              & 235.00                    & \textbf{257.36}              & 337.11                    & 134.79                       & \textbf{87.99}            \\
                                                 & \textit{Cosine}      & 131.39                       & \textbf{110.01}           & \textbf{206.00}              & 233.00                    & \textbf{257.73}              & 334.86                    & 127.63                       & \textbf{87.96}            \\
                                                 & \textit{Dice}        & 130.52                       & \textbf{110.94}           & \textbf{206.00}              & 233.00                    & \textbf{130.35}              & 335.49                    & 134.36                       & \textbf{88.19}            \\
                                                 & \textit{Euclidean}   & 131.93                       & \textbf{111.38}           & \textbf{206.00}              & 233.00                    & \textbf{258.38}              & 336.63                    & 134.75                       & \textbf{87.77}            \\
                                                 & \textit{Hellinger}   & 129.79                       & \textbf{110.82}           & \textbf{205.00}              & 232.00                    & \textbf{258.22}              & 334.80                    & 134.11                       & \textbf{87.83}            \\
                                                 & \textit{Jaccard}     & 130.51                       & \textbf{110.67}           & \textbf{206.00}              & 233.00                    & \textbf{258.24}              & 336.01                    & 134.55                       & \textbf{87.73}            \\
                                                 & \textit{Russel-Rao}  & 130.35                       & \textbf{109.68}           & \textbf{206.00}              & 232.00                    & \textbf{257.58}              & 332.93                    & 134.31                       & \textbf{87.94}            \\ \midrule
\multirow{7}{*}{\STAB{\rotatebox[origin=c]{90}{Non-Trivial Metrics}}} & Canberra             & 3014.34                      & \textbf{268.11}           & 4027.00                      & \textbf{598.00}           & 4164.98                      & \textbf{819.80}           & 505.71                       & \textbf{102.79}           \\
                                                 & Chebyshev            & 1621.00                      & \textbf{336.05}           & 3907.00                      & \textbf{546.00}           & 2709.30                      & \textbf{1072.35}          & 253.00                       & \textbf{146.41}           \\
                                                 & Hamming              & 1635.30                      & \textbf{229.59}           & 3902.00                      & \textbf{481.00}           & 2724.86                      & \textbf{728.05}           & 258.27                       & \textbf{97.65}            \\
                                                 & Jensen-Shannon       & 7187.27                      & \textbf{415.12}           & 4257.00                      & \textbf{1052.00}          & 10869.32                     & \textbf{1331.37}          & 1248.83                      & \textbf{142.96}           \\
                                                 & KL Divergence        & 5013.65                      & \textbf{170.06}           & 4117.00                      & \textbf{409.00}           & 7099.08                      & \textbf{525.32}           & 753.56                       & \textbf{87.72}            \\
                                                 & Manhattan            & 1632.05                      & \textbf{227.98}           & 3904.00                      & \textbf{477.00}           & 2699.91                      & \textbf{715.78}           & 254.69                       & \textbf{98.05}            \\
                                                 & Minkowski            & 1632.05                      & \textbf{367.17}           & 4051.00                      & \textbf{838.00}           & 5855.79                      & \textbf{1161.31}          & 646.71                       & \textbf{129.47}           \\ \bottomrule
\end{tabular}
\end{table*}

\subsection{Runtime Performance}

To get an idea of how each supported distance performed on data of different shapes and degree distributions, we benchmarked all of the supported distances for each of the datasets, even though some of them may provide irrelevant geometries in practice. Benchmarks were performed on a DGX1 containing dual 20-core Intel Xeon ES-2698 CPUs (80 total threads) at 2.20GHZ and a Volta V100 GPU running CUDA 11.0 for both the driver and toolkit. Each benchmark performs a k-nearest neighbors query to test our primitives end-to-end  and allow scaling to datasets where the dense pairwise distance matrix may not otherwise fit in the memory of the GPU. We used the brute-force \textit{NearestNeighbors} estimator from RAPIDS cuML for the GPU benchmarks since it makes direct use of our primitive. We used Scikit-learn's corresponding brute-force \textit{NearestNeighbors} estimator as a CPU baseline and configured it to use all the available CPU cores. Each experiment trains the \textit{NearestNeighbors} estimator on the entire dataset and then queries the entire dataset, timing only the query. Compared to the CPU, we observed an average of $28.78\times$ speedup for the dot-product-based distances and $29.17\times$ speedup for the distances which require the non-annihilating product monoid.

\begin{figure}[!h]
 \begin{minipage}{0.991\columnwidth}
  \centering
\begin{minted}[breaklines]{python}
from cuml.neighbors import NearestNeighbors
nn = NearestNeighbors().fit(X)
dists, inds = nn.kneighbors(X)
\end{minted}
 \end{minipage}
 
 \begin{minipage}{0.991\columnwidth}
  \centering
\begin{minted}[breaklines]{python}
from cuml.metrics import pairwise_distances
dists = pairwise_distances(X, metric='cosine')
\end{minted}
 \end{minipage}
 \caption{Excluding data loading and logging, all the code needed to perform the same GPU accelerated sparse distance calculations done in this paper are contained within these two snippets. Top shows k-NN search, bottom all pairwise distance matrix construction. These are the APIs that most would use.}
  \label{lst:representation_examples}
\end{figure}

From the strategies described in Section \ref{sec:gpu_acceleration}, we benchmarked our best performing approach, the Load-balanced Hybrid COO+CSR SPMV described in \autoref{sec:load_balanced_semiring}, using the hash table strategy to sparsify the vector in shared memory.

\begin{figure}[!h]
\begin{minted}[fontsize=\small,breaklines]{c++}
#include <raft/sparse/distance/coo_spmv.cuh>
#include <raft/sparse/distance/operators.h>

using namespace raft::sparse::distance

distances_config_t<int, float> conf;

// Use conf to set input data arguments...

balanced_coo_pairwise_generalized_spmv(
    out_dists, conf, coo_rows_a, 
    AbsDiff(), Sum(), AtomicSum());

balanced_coo_pairwise_generalized_spmv_rev(
    out_dists, conf, coo_rows_b, 
    AbsDiff(), Sum(), AtomicSum());
\end{minted}
\caption{The C++ API can be used to construct new semirings. Dot-product-based semirings only need invoke the first function while NAMMs can be constructed by invoking both. While the Python API is part of the RAPIDS cuML project, the C++ API is provided by the RAFT project (http://github.com/rapidsai/raft). RAFT is a header only library that contains fundamental algorithms and primitives for data science, graph, and machine learning applications.}
\end{figure}

As evidenced in table \ref{tbl:gpu_results_movielens},
our implementation consistently outperforms the CPU. We also outperform the baseline, cuSPARSE, for the distances that it supports in two out of the four datasets. In addition to maintaining comparable performance in the remaining two datasets, our design is also flexible enough to provide distances which require the NAMM outlined in \autoref{subsec:semirings} while using less memory. As mentioned in \autoref{sec:related_work}, it is not uncommon to see different sparse implementations performing better on some datasets than others \cite{sedaghati2015characterizing} and the flexibility of our implementation, as well as our well-defined set of rules for supporting a wide array of distances, will allow us to continue optimizing our execution strategies to support patterns that we find frequently occurring across different sparse datasets.

\subsection{Memory Footprint}

The density of the dot product matrix that is returned from the cuSPARSE \textit{csrgemm()} is fully dependent upon the dataset. Because 2 arrays, each of size $nnz$, are required to represent the cuSPARSE output in CSR format, a density of 50\% would require the same amount of space as the full dense pairwise distance matrix. A density of 100\% requires 2x the amount of space as the dense pairwise distance matrix. In addition, since the output still needs to be converted to a dense format, this requires an additional allocation of the dense pairwise distance matrix in a space of contiguous memory locations even if the cuSPARSE output was 99.9\% dense. We found the density of the cuSPARSE output to be at least $57\%$ on average across the batches for Movielens, $\>98\%$ for NY Times BoW and was fully dense in scRNA. The SEC Edgar datasets had the highest variance in density from batch-to-batch and were significantly different between n-gram sizes. The unigram and bigram dataset ranged from $5\%$ to $25\%$ output density, for example, while trigrams ranged from $24\%$ to $43\%$. 

This provides further evidence of the subtle but important differences between the types of data we expect to encounter in neighborhood methods, however even more evident is that the matrix resulting from computing the dot product semiring over the square connectivities graphs used in other sparse matrix multiplication research \cite{williams2007optimization,bell2008efficient} is extremely sparse. In addition to the output memory, cuSPARSE required an internal temporary workspace in device memory with anywhere from 300mb to 550mb of additional memory per batch while our dot product semiring required a workspace buffer of size $nnz(B)$ per batch. Strangely, the size of this temporary workspace seemed almost identical even when computed on the square connectivities graphs mentioned above.

\section{Related Work}

\label{sec:related_work}



\subsection{Sparse matrix multiplication}

The task of efficient and performant sparse matrix multiplication is an active area of research, with implementations spanning the spectrum of scientific computing. In high performance computing environments, these solutions are designed around both hardware and software constraints \cite{jeon2020biqgemm,guo2020accelerating,gale2020sparse,gray2017gpu, bell2008efficient}, often making use of specialized hardware capabilities and optimizing for specific sparsity patterns, an unfortunate side-effect that can reduce their potential for reuse. What complicates this further are the number of different optimized variants of sparse matrix multiplication available in open source libraries, each using different concurrency patterns and available memory to provide speedups based on either supported sparse formats or the assumed density of either the inputs or the outputs \cite{sedaghati2015characterizing,mattson2013standards}. We have compareda against the seminal cuSPARSE~\cite{naumov2010cusparse} that is highly optimized for sparse dot product based k-nearest neighbors \cite{zhou2018gpu}, and found our approach is faster or competitive in all cases, but is not limited to dot product based measures.

Better able to make use of critical optimizations inherent in their dense counterparts, block compressed sparse formats have become widely popular for representing sparse data \cite{zachariadis2020accelerating}, in part because they can improve load balancing by grouping nonzeros into fixed-sized tiles and scheduling the tiles more uniformly across the processing cores. Enabling sparse formats to be processed more similar to their dense counterparts allows the use of specialized hardware optimizations such as tensor cores. While we do hope to someday support block-sparse formats, it is most often assumed that users will be calling code that invokes our primitive with matrices in the standard compressed sparse row (CSR) format \cite{williams2007optimization} 
and so a conversion would be necessary in order to use a blocked format.




\subsection{Semirings}

Consolidating seemingly disparate concepts into a lightweight, terse, and abstract set of building-blocks can increase flexibility and promote reuse \cite{izbicki2013algebraic}. This especially benefits fields which require non-trivial and highly-optimized implementations where the design complexities and costs are high, the basic linear-algebra subroutines (BLAS) API and GPU-accelerated computing being common examples. Semirings provide the efficiency and flexibility to enable algorithms in which the representation and assumptions of the typical BLAS API for dense linear algebra comes up short \cite{mattson2013standards}. NIST published a sparse BLAS standard back in 2001 \cite{duff2002overview} and cuSPARSE is one of the most sophisticated implementations of the sparse BLAS standard that has been built on the GPU, however as mentioned above, its multiplication routines fix the inner product to the dot product. GraphBLAS~\cite{Davis2018} provides a set of primitives, along with an API, for using semiring algebras to implement graph algorithms. The GraphBLAST~\cite{yang2019graphblast} and SuiteSparse~\cite{davis2019algorithm} libraries provide implementations of the GraphBLAS that also include GPU-accelerated primitives.


The use of semirings in graph theory dates back to the early 1970s \cite{ratti1971graphs}, when "good old-fashioned artificial intelligence", or \textit{Symbolic AI}, was a dominant paradigm in research. Semirings have also been used for some time to implement more modern machine learning methods \cite{belle2020semiring}, with the more recent invention of semiring programming attempting to further consolidate these concepts under a single framework and set of symbolic routines. Semirings can be a useful building-block for linear models ~\cite{jananthan2017linear}, probabilistic models, such as Bayesian networks~\cite{wachter2007optimizing} and the use of Tropical semiring in Markov networks~\cite{ilic2011entropy}. The Tropical semiring is also being used to implement sparse non-negative matrix factorizations \cite{omanov2020data}. 


\subsection{Neighborhood Methods}


Our work is positioned to have an impact on numerous down-stream tasks that often depend on sparse nearest-neighbor retrieval. This includes classic Information Retrieval problems where such methods are still highly competitive or preferred \cite{Mitra2018,Li2016c,Soboroff:2018:MRE:3269206.3271719,10.1145/3086701,Bouthillier2021}. Dimensional reduction approaches like t-SNE~\cite{Maaten2008} and UMAP~\cite{McInnes2018} that lack sparse input support on GPUs without our method~\cite{nolet2020bringing}. ML models based on the kernel trick, such as Guassian Process \cite{lawrence2009non} also stand to benefit. The breadth and frequency of nearest neighbor methods on high dimensional data make our work relevant to an especially wide class of practioners.

\section{Conclusion} 
\label{sec:conclusion}

In this paper, we demonstrated a flexible sparse pairwise distance primitive that is able to collectively support, to our knowledge, a larger assortment of widely-used distance measures than any other package on the GPU. We consolidated the design of these distance measures using a couple minor enhancements to the rules of classical semirings, which are traditionally used to implement graph algorithms, and we discussed the impact of our primitive as a core building block of many important neighborhood methods for machine learning and data mining. Finally, we provided a novel implementation as an example of how these semirings can be implemented on the GPU with a lower memory footprint and performance comparable to, or better than, the current state of the art.




\bibliographystyle{ACM-Reference-Format}
\bibliography{references}

\clearpage
\appendix

\section{Appendix}

\subsection{Deriving Distances With Semirings}
\label{deriving_distances_with_semirings}

All of the distances in this paper can be categorized into one of two groups- those which can be computed using the dot product and vector norms and those which cannot. The non-annhiliating multiplicative monoid (NAMM) is used for the latter group, which requires exhaustive computation over the union of non-zeros from each input. The following example derives the semiring for the Manhattan distance, demonstrating why the dot-product cannot be used.

Let vector $a = [1, 0, 1]$ and $b = [0, 1, 0]$

We can compute the L1 distance between these two vectors by taking the sum of the absolute value of their differences:

\begin{align}
\sum(|a-b|) & = \\
\sum([|1 - 0|, |0 - 1|, |1 - 0|]) & = \\ 
\sum([1, 1, 1]) & = 3 
\end{align}

Semiring standards such as GraphBLAS, for example, often make use of the detail that the multiplicative annihilator is equal to the additive identity. If we follow this detail in our example, we end up with the following result (if any side is 0, the arithmetic evaluates to 0):

\begin{align}
\sum(|a-b|) & = \\
\sum([|1-0|, |0-1|, |1-0|]) & = \\
\sum([0, 0, 0]) & = 0
\end{align}

What we need here instead is for the multiplicative identity to be non-annihilating, such that it equals the additive identity, so that the difference in our example behaves like an XOR, evaluating to the other side when either side is zero and evaluating to 0 only in the case where both sides have the same value. For example, 

$|1-0| = 1$
$|0-1| = 1$
$|0-0| = 0$
$|1-1| = 0$

Now let’s perform a sparse-matrix sparse-vector multiply where $A = [[1, 0, 1]]$ and $b = [0, 1 , 1]$

We can parallelize this by evaluating the semiring of b over each row vector of A independently, iterating through the nonzero columns from each vector in A and fetching or looking up the corresponding column from b (if it is nonzero). With the standard dot-product semiring, which annihilates multiplicatively over the additive identity, we only need to consider the intersection of columns where both sides are nonzero– column 3 in this example.

Removing the multiplicative annihilator results in the need to consider the union of non-zero columns, and so all columns need to be considered in this example. However if only the nonzero columns in the vectors of A are visited, the nonzero columns in b, which are zero in A, will be missed. 

Recall that we can decompose a full union across all nonzero columns into a union of 
the symmetric difference between nonzero columns of A and b (that is, all columns which are nonzero in A and zero in b), 
the intersection between nonzero columns of A and b (where both sides are nonzero), and 
the symmetric difference between the nonzero columns of b and A (that is, all columns which are nonzero in b and zero in A). 

A spmv will often compute the intersection between the nonzero columns of A and b and the symmetric difference between nonzero columns of A and b will be computed only as a side-effect. In order to compute the union between the nonzero columns of A and b, the symmetric difference between the nonzero columns of b and A still needs to be computed. We compute this with a second pass of the spmv by flipping the inputs to the spmv and ignoring the intersecting columns in the second pass.

\end{document}